\pdfoutput=1

\documentclass[11pt]{article}

\usepackage{acl}

\usepackage{times}
\usepackage{latexsym}

\usepackage[T1]{fontenc}

\usepackage[utf8]{inputenc}

\usepackage{microtype}
\usepackage{enumerate}
\usepackage{enumitem}
\usepackage{multirow}
\usepackage{graphicx}
\usepackage{array}
\usepackage{booktabs}
\usepackage{newtxmath}

\usepackage{inconsolata}
\usepackage{amsmath,amsfonts,bm}

\newcommand{\Na}{({\em a})~}
\newcommand{\Nb}{({\em b})~}
\newcommand{\Nc}{({\em c})~}

\makeatletter   
\newcommand{\sveryshortarrow}[1][3pt]{\mathrel{%
    \vcenter{\hbox{\rule[-.5\fontdimen8\scriptfont3]
               {\scriptratio\dimexpr#1\relax}{\fontdimen8\scriptfont3}}}%
   \mkern-4mu\hbox{\let\f@size\sf@size\usefont{U}{lasy}{m}{n}\symbol{41}}}}
\makeatother

\def\eqref#1{equation~\ref{#1}}

\def\1{\bm{1}}

\def\m1{{\bm{1}}}

\DeclareMathAlphabet{\mathsfit}{\encodingdefault}{\sfdefault}{m}{sl}
\SetMathAlphabet{\mathsfit}{bold}{\encodingdefault}{\sfdefault}{bx}{n}

\def\gT{{\mathcal{T}}}

\def\sT{{\mathbb{T}}}

\usepackage[nameinlink]{cleveref}
\crefformat{section}{\S#2#1#3} 
\crefname{algorithm}{Alg.}{Algs.}
\crefformat{subsection}{\S#2#1#3}
\Crefname{equation}{Eq.}{Eqs.}
\Crefname{figure}{Fig.}{Figs.}
\newcommand{\esco}{\textsc{ESCO}}

\newcommand{\led}{\textsc{led}}
\newcommand{\ie}{{\em i.e.,}}
\newcommand{\eg}{{\em e.g.,}}
\newcommand{\red}[1]{\textcolor{purple}{#1}}
\newcommand{\blue}[1]{\textcolor{blue}{#1}}

\definecolor{applegreen}{rgb}{0.56, 0.8, 0.25}

\title{Lifelong Event Detection with 

Embedding Space Separation and Compaction}

\author{Chengwei Qin$^\dagger$, Ruirui Chen$^\vardiamondsuit$, Ruochen Zhao$^\dagger$, Wenhan Xia$^\clubsuit$, Shafiq Joty$^\dagger$$^\spadesuit$\\
$^\dagger$Nanyang Technological University $^\clubsuit$Princeton University $^\spadesuit$Salesforce Research\\
$^\vardiamondsuit$Institute of High Performance Computing (IHPC),\\
Agency for Science, Technology and Research (A*STAR)\\
1 Fusionopolis Way, \#16-16 Connexis, Singapore 138632, Republic of Singapore\\
\texttt{\{chengwei003@e.ntu, ruochen002@e.ntu, srjoty@ntu \}.edu.sg}\\ \texttt{chen\_ruirui@ihpc.a-star.edu.sg}, \texttt{wxia@princeton.edu}
}

\begin{document}
\maketitle
\begin{abstract}
To mitigate forgetting, existing lifelong event detection methods typically maintain a memory module and replay the stored memory data during the learning of a new task. However, the simple combination of memory data and new-task samples can still result in substantial forgetting of previously acquired knowledge, which may occur due to the potential overlap between the feature distribution of new data and the previously learned embedding space. Moreover, the model suffers from overfitting on the few memory samples rather than effectively remembering learned patterns. To address the challenges of forgetting and overfitting, we propose a novel method based on embedding space separation and compaction. Our method alleviates forgetting of previously learned tasks by forcing the feature distribution of new data away from the previous embedding space. It also mitigates overfitting by a memory calibration mechanism that encourages memory data to be close to its prototype to enhance intra-class compactness. In addition, the learnable parameters of the new task are initialized by drawing upon acquired knowledge from the previously learned task to facilitate forward knowledge transfer. With extensive experiments, we demonstrate that our method can significantly outperform previous state-of-the-art approaches.
\end{abstract}

\section{Introduction} \label{sec:intro}

Event detection (ED) aims to detect the event type of trigger words in a given sentence, \eg\, extracting the event type \emph{injure} from the trigger word \emph{scalded} in text ``\emph{He was scalded by hot water}''. Traditional ED methods typically consider a fixed pre-defined set of event types ~\citep{chen-etal-2015-event,nguyen-etal-2016-joint-event,huang-ji-2020-semi, chen2024large}. However, as the environment and data distributions change in real scenarios, the model might face challenges in handling rapidly emerging new types \citep{samuel}.

A more practical setting is lifelong event detection or \led\ \citep{cao-etal-2020-incremental}, where the model learns event knowledge from a sequence of tasks with different sets of event types. In \led, the model is expected to retain and accumulate knowledge when learning new tasks, which is challenging due to \emph{catastrophic forgetting} \citep{mccloskey1989catastrophic} of previously acquired knowledge. Existing methods \citep{cao-etal-2020-incremental,yu-etal-2021-lifelong} for mitigating forgetting in \led\ typically maintain a memory that saves a few key samples of previous tasks, which are then combined with new data for training. Recently, \citet{liu-etal-2022-incremental} introduce Episodic Memory Prompts (EMP) that leverages soft prompts to remember learned event types, achieving state-of-the-art performance on \led.

Despite its effectiveness, EMP has two key limitations. First, simply combining new data and memory samples for training can still result in forgetting as the feature distribution of new data might overlap with the previously learned embedding space (see \Cref{sec:overlap_distribution}). Second, it may overfit on a few memory samples after frequent replays rather than effectively retaining learned patterns. 

To address the above limitations of EMP, in this paper, we introduce a novel method based on Embedding space Separation and COmpaction (\esco) for \led. In particular, we propose a margin-based loss that forces the feature distribution of new event types away from the learned embedding space to alleviate \emph{forgetting}. Inspired by \citet{han-etal-2020-continual}, we introduce a memory calibration mechanism to encourage memory data to be close to its prototype to avoid \emph{overfitting} on the few memory samples. In addition, the learnable parameters of the new task are initialized using those of the previously learned task to facilitate \emph{forward knowledge transfer}, which is as important for lifelong learning as preventing forgetting \citep{ke2020continual,qin2023lifelong,qin-etal-2023-learning}. The empirical results show that our method significantly outperforms previous state-of-the-art approaches. In summary, our main contributions are: 

\begin{itemize}[leftmargin=*,topsep=2pt,itemsep=2pt,parsep=0pt]
    \item We propose \esco, a novel method based on embedding space separation and compaction to mitigate forgetting and overfitting in \led.
    \item With extensive experiments and analysis, we demonstrate the effectiveness of our method compared to existing ones. 
\end{itemize}
\section{Problem Formulation}

\led\ involves learning from a stream of event detection tasks $\sT = (\gT^1,  \ldots, \gT^n)$, where each task $\mathcal T^k$ has its own training set $\mathcal D_{\text{train}}^k$, validation set $\mathcal D_{\text{valid}}^k$, and test set $\mathcal D_{\text{test}}^k$. For every input text $x^i$ in $\mathcal D^k$, it contains a set of target spans $\{ \overline x^i_t \}$ and their corresponding labels $y^i_t$ which belong to the event type set $\mathcal{C}^k$ of task $\mathcal{T}^k$. Note that the event type sets of different tasks are non-overlapping.

After the training on $\mathcal D_{\text{train}}^k$, the model is expected to perform well on all the $k$ tasks that it has learned and will be evaluated on the combined test set $\hat{\mathcal{D}}_{\text{test}}^k = \cup_{i=1}^{k} \mathcal D_{\text{test}}^i$ consisting of all known event types $\hat{\mathcal{C}}^k = \cup_{i=1}^{k} \mathcal{C}^i$. During the learning, a memory module $\mathcal M$ which stores a few key samples of previous tasks is maintained to overcome the forgetting problem. 
\section{Embedding Space Separation and Compaction} \label{sec:methodology}
When learning a new task $\mathcal T^k$, following \citet{liu-etal-2022-incremental}, we first initialize a set of soft prompts $\mathcal{P}^k = \{ p^k_1,...,p^k_{|\mathcal{C}^k|} \}$ where $\mathcal{C}^k$ is the event type set of $\mathcal T^k$. The accumulated prompts $\mathcal{Q}^k = [\mathcal{P}^1,...,\mathcal{P}^k]$ until time step $k$ are then combined with the input text $x^i$ to obtain the contextual representations using a frozen BERT \citep{devlin-etal-2019-bert}:
\begin{equation}
\begin{aligned}
 [\textbf{x}^i, \textbf{Q}^k] = \text{BERT} ([x^i, \mathcal{Q}^k])
\end{aligned}
\end{equation}
where $\textbf{x}^i$ and $\textbf{Q}^k$ are the representations of $x^i$ and $\mathcal{Q}^k$, respectively. To facilitate \emph{forward knowledge transfer} \citep{qin2022lfpt}, we initialize soft prompts $\mathcal{P}^k$ of the new task using learned prompts $\mathcal{P}^{k-1}$ of the previous task. For the first task $\mathcal T^1$, we initialize each event type prompt $p^1_i$ in $\mathcal{P}^1$ using its corresponding name.

To predict the event type of the span $\overline x^i_t$, we concatenate the representations corresponding to the start and end token and obtain the logits over all learned types through a feed-forward network (FFN) as well as a linear layer:
\begin{equation}
\begin{aligned}
 Z^i_t = \text{Linear} ({\text{FFN} ([\textbf{x}^i_m, \textbf{x}^i_n])})
\end{aligned}
\end{equation}
where $\overline{\textbf{x}}^i_t = \text{FFN} ([\textbf{x}^i_m, \textbf{x}^i_n])$ is the span representation, $m$ and $n$ denote the start and end index of the span, respectively. 
Following \citet{liu-etal-2022-incremental}, to entangle span representations with soft prompts, the probability distribution over all prompts is calculated as $Z_q = \text{FFN} (\textbf{Q}^k) \cdot \overline{\textbf{x}}^i_t$, where $\cdot$ is the inner product. $Z_q$ is then combined with $Z^i_t$ to optimize the cross entropy loss:
\begin{equation}
\begin{aligned}
 \mathcal{L}_{\text{new}} = - \sum_{(\overline x^i_t,y^i_t) \in \mathcal D_{\text{train}}^k } \text{CE} (Z^i_t + Z_q, y^i_t)
\end{aligned}
\label{eq:entangled_loss}
\end{equation}

After learning the previous task $\mathcal T^{k-1}$, we select the top-$l$ most informative training examples for each event type in $\mathcal{C}^{k-1}$ using the herding algorithm \citep{welling2009herding}, which are then saved in the memory module $\mathcal M$ for replay to mitigate forgetting. Similar as Eq. \ref{eq:entangled_loss}, the training objective for memory replay when learning $\mathcal T^k$ is:

\begin{equation}
\begin{aligned}
 \mathcal{L}_{\text{mem}} = - \sum_{(\overline x^i_t,y^i_t) \in \mathcal M } \text{CE} (Z^i_t + Z_q, y^i_t)
\end{aligned}
\end{equation}

However, the simple combination of $\mathcal{L}_{\text{new}}$ and $\mathcal{L}_{\text{mem}}$ can still result in substantial forgetting of acquired knowledge due to the potential overlap between the feature distribution of new event types and the previously learned embedding space (see \Cref{sec:overlap_distribution}). To ensure that the new feature distribution is away from the learned embedding space, we design a \emph{margin-based} loss, which decreases the similarity scores between new samples and prototypes (see Eq. \ref{eq:prototype} for the calculation of prototypes) of learned event types:

\begin{equation}
\small
\begin{aligned}
    \mathcal{L}_{\text{sim}} = \sum_{(\overline x^i_t,y^i_t) \in \mathcal D_{\text{train}}^k } \sum_{\textbf{e}_i \in \mathcal{E}^{k-1}} \max ( 0,  g(\overline{\textbf{x}}^i_t,\textbf{e}_i) - m_1)
\end{aligned}
\end{equation}
where $\mathcal{E}^{k-1}$ is the prototype set of previous $k-1$ tasks, $g(,)$ is the similarity function (cosine similarity) and $m_1$ is the margin for $\mathcal{L}_{\text{sim}}$. Note that $\mathcal{L}_{\text{sim}}$ is different from metric learning or contrastive learning \citep{qin-joty-2022-continual} which typically considers both positive and negative pairs. $\mathcal{L}_{\text{sim}}$ only includes negative pairs while ignoring positive ones as our goal in designing $\mathcal{L}_{\text{sim}}$ is to \emph{separate} the new feature distribution and the learned embedding space.

As the size of memory $\mathcal M$ is typically small, the model is prone to overfit on the few memory samples after frequent replays, making learned distributions distorted. To effectively recover from distorted learned distributions, we introduce a \emph{memory calibration} mechanism inspired by \citet{han-etal-2020-continual}. Specifically, for each memory sample in $\mathcal M$, we encourage it to be close to its corresponding prototype to improve the intra-class compactness of learned distributions. More formally,

\begin{equation}
\small
\begin{aligned}
    \mathcal{L}_{\text{cal}} = -\sum_{(\overline x^i_t,y^i_t) \in \mathcal M } \log \frac{\exp g(\overline{\textbf{x}}^i_t,\textbf{e}_l)}{\sum_{j=1}^{|\mathcal{E}^{k-1}|} \exp g(\overline{\textbf{x}}^i_t,\textbf{e}_j)}
\end{aligned}
\end{equation}
where $\textbf{e}_l$ is the prototype of $y^i_t$. The \emph{total} loss for learning on $\mathcal{T}^k$ is defined as:
\begin{equation}
    \mathcal{L}_{\text{total}} = \mathcal{L}_{\text{new}} + \lambda_1 \mathcal{L}_{\text{sim}} + \lambda_2 (\mathcal{L}_{\text{mem}} + \mathcal{L}_{\text{cal}})
\end{equation}
where $\lambda_1$ and $\lambda_2$ are loss weights. 

After learning $\mathcal{T}^k$ and selecting memory data for $\mathcal{T}^k$, we use the memory $\mathcal M$ to calculate prototypes of all learned event types in $\mathcal{C}^k$. Specifically, for each event type $e_j$ in $\mathcal{C}^k$, we obtain its prototype $\textbf{e}_j$ by averaging the span representations of all samples labeled as $e_j$ in $\mathcal M$ as follows:
\begin{equation}
    \textbf{e}_j =  \frac{1}{|\mathcal M_{e_j}|} \sum_{(\overline x^i_t,y^i_t) \in \mathcal M_{e_j}} \overline{\textbf{x}}^i_t
    \label{eq:prototype}
\end{equation}
where $\mathcal M_{e_j} = \{ (\overline x^i_t,y^i_t) | (\overline x^i_t,y^i_t) \in \mathcal M, y^i_t = e_j \}$.

\section{Experiment} \label{sec:exp}

\subsection{Experimental Setup}

We conduct experiments on two representative event detection datasets in our work: ACE05 \citep{doddington-etal-2004-automatic} and MAVEN \citep{wang-etal-2020-maven}. Following \citet{liu-etal-2022-incremental}, we divide each dataset into a sequence of 5 tasks with non-overlapping event type sets to form a \emph{class-incremental} setting. After learning $\mathcal T^k$, the model is evaluated on the combined test set $\hat{\mathcal{D}}_{\text{test}}^k = \cup_{i=1}^{k} \mathcal D_{\text{test}}^i$ of all seen tasks. As the task order might influence the model performance, we run experiments for each dataset 5 times with different task order permutations and report the average results. More details of the experimental setup are in \Cref{sec:exp_details}.

\subsection{Methods Compared}
We compare our approach with the following methods: (1) \textbf{Fine-tuning} tunes the model only on new data without memory; (2) \textbf{BiC} \citep{wu2019large} introduces a bias correction layer to improve lifelong learning performance; (3) \textbf{KCN} \citep{cao-etal-2020-incremental} designs prototype enhanced retrospection and hierarchical distillation to alleviate semantic ambiguity and class imbalance; (4) \textbf{KT} \citep{yu-etal-2021-lifelong} proposes to transfer knowledge between related types; (5) \textbf{EMP} \citep{liu-etal-2022-incremental} leverages type-specific soft prompts to remember learned event knowledge; and (6) \textbf{Multi-task learning (MTL)} simultaneously trains the model on all data, serving as the \emph{upper bound} in \led.

\subsection{Main Results}

\begin{table*}[!t]
	\centering
	\resizebox{1.00\linewidth}{!}
	{
	\begin{tabular}{l | ccccc | ccccc}
    \toprule
    & \multicolumn{5}{c}{MAVEN}  & \multicolumn{5}{c}{ACE05} \\
    \midrule
    Task index &1 &2 & 3 & 4 & 5 & 1 &2 & 3 & 4 & 5  \\ 
    \midrule
    Fine-tuning & 63.51 & 39.99 & 33.36 & 23.83 & 22.69 & 58.30 & 43.96 & 38.02 & 21.53 & 25.71 \\
    BiC & 63.51 & 46.69 & 39.15 & 31.69 & 30.47 & 58.30 & 45.73 & 43.28 & 35.70 & 30.80 \\
    KCN & 63.51 & 51.17 & 46.80 & 38.72 & 38.58 & 58.30 & 54.71 & 52.88 & 44.93 & 41.10 \\
    KT  & 63.51 & 52.36 & 47.24 & 39.51 & 39.34 & 58.30 & 55.41 & 53.95 & 45.00 & 42.62 \\
    EMP & \textbf{67.50} & 59.67 & 58.03 & 54.80 & 54.39 & \textbf{58.35} & 50.03 & 54.91 & 47.78 & 47.19 \\ 
    \midrule
    \esco & \textbf{67.50} & \textbf{61.37} & \textbf{60.65} & \textbf{57.43} & \textbf{57.35} & \textbf{58.35} & \textbf{57.42} & \textbf{57.63} & \textbf{53.64} & \textbf{55.20} \\ 
    \midrule
    MTL & --- & --- & --- & --- & 68.42 & --- & --- & --- & --- & 67.22 \\
    \bottomrule
    \end{tabular}
    } 
    \caption{F1 score ($\%$) of different methods at every time step on two datasets. `MTL' stands for `multi-task learning'. \esco\ is significantly better than EMP with $p$-value $<0.05$ (paired t-test). We report results with variance and detailed results for different task orders in \Cref{sec:results_with_variance} and \Cref{sec:results_different_oders}, respectively. 
    }
    \label{tab:main_result}
\end{table*}

\begin{figure}[t]
    \centering
    \includegraphics[width=0.48\textwidth]{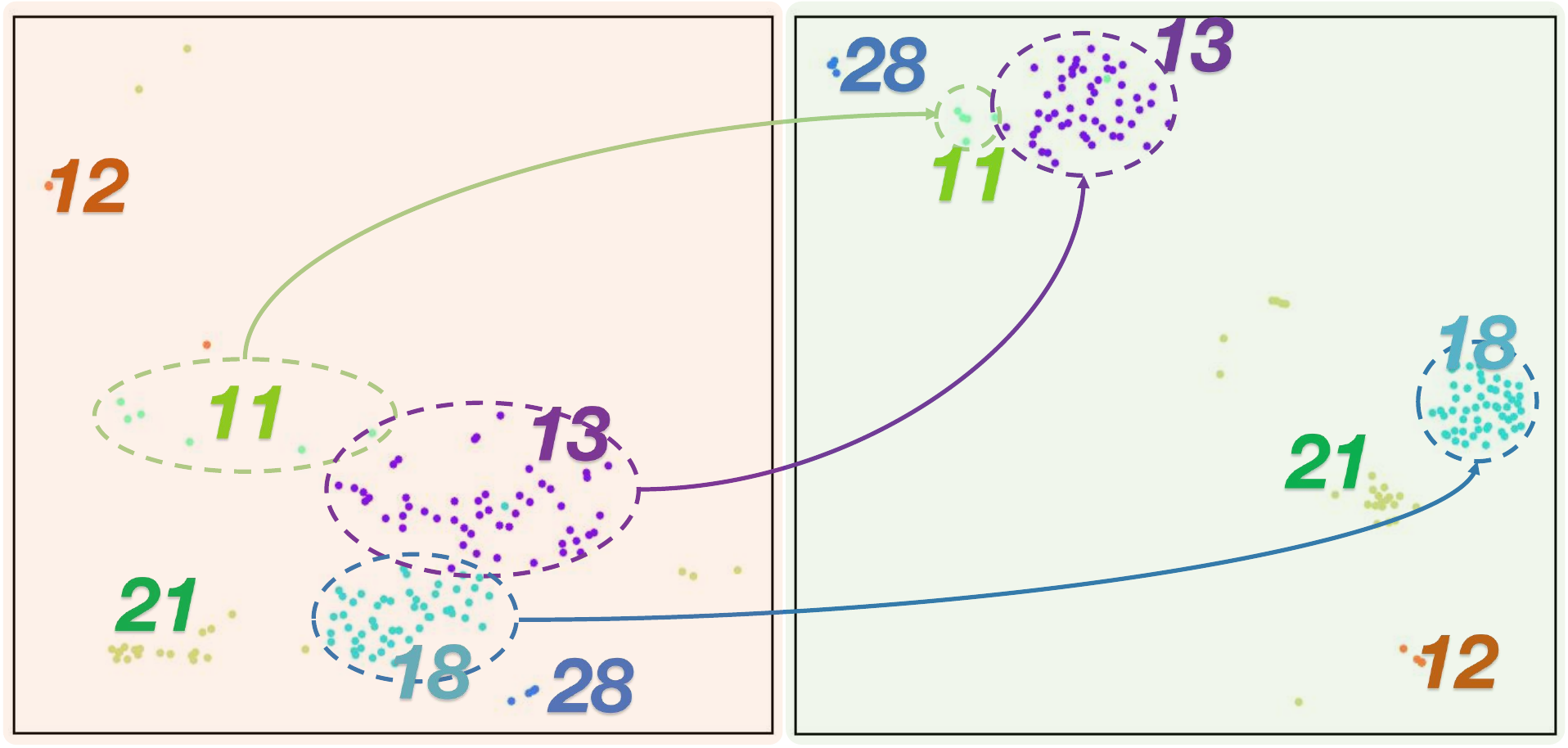}
    \caption{Comparison between the embedding spaces of EMP (left) and \esco\ (right). Colors represent different event types with numbers being the event indexes. Compared with EMP, \esco\ shows larger inter-class distances, \eg\ the distance between 13 and 18, and better intra-class compactness (circled regions).}
    \label{fig:diff_vis}
\end{figure}

We report the F1 score of different methods at each time step in \Cref{tab:main_result}. From the results, we can observe that \esco\ significantly outperforms previous baselines on both datasets, demonstrating its superiority. Simply fine-tuning the model on new data without memory replay results in poor performance due to severe forgetting of learned knowledge. Although BiC, KCN and KT could alleviate forgetting to some extent, there is still a large performance drop after learning all tasks. EMP achieves better performance because the type-specific soft prompts help retain previously acquired knowledge. However, it does not necessarily ensure large distances among feature distributions of different event types, and easily overfits on the memory samples. Our proposed \esco\ outperforms EMP by a large margin through embedding space separation and compaction. To verify its effectiveness, we visualize the embedding spaces of EMP and \esco\ on ACE05 in \Cref{fig:diff_vis}. Specifically, we randomly select 6 event types from different learning stages and visualize their test data using t-SNE \citep{van2008visualizing}. The comparison demonstrates that \esco\ could achieve larger inter-class distances and better intra-class compactness in the embedding space.

\subsection{Ablation Study}

To analyze the contribution of different components of \esco, we conduct several ablations. Specifically, we investigate three variants of \esco: \Na without the margin-based loss (\emph{w.o.} margin), \Nb removing the memory calibration mechanism (\emph{w.o.} calibration), and \Nc without forward knowledge transfer (FKT) (\emph{w.o.} FKT). The results of different ablations after learning all tasks are reported in \Cref{table:ablation-brief}. We can see that all components contribute to overall performance. The margin-based loss yields about \textbf{1.25}$\%$ performance boost as it can bring feature distribution of new data away from the learned embedding space. The memory calibration mechanism improves the F1 score by \textbf{1.47}$\%$, which demonstrates the necessity of improving intra-class compactness of learned distributions. The adoption of forward knowledge transfer leads to \textbf{0.80}$\%$ improvement, indicating that it can indeed transfer useful learned knowledge to facilitate the learning of new tasks.

\subsection{Further Analysis} \label{subsec:further_analysis}

\paragraph{Quantify Knowledge Transfer.} Following \citet{lopez2017gradient}, we report the backward transfer (BWT) and forward transfer (FWT) of EMP and \esco\ after learning all tasks in \Cref{tab:knowledge_transfer}. From the comparison, we can observe that ESCO outperforms EMP by a large margin in terms of BWT and FWT on both datasets, demonstrating its effectiveness.

In addition, we show results of a different backbone model (RoBERTa \citep{liu2019roberta}), the effect of memory size, results of different memory sample selection methods, the comparison with the contrastive loss in \citet{qin-joty-2022-continual}, and a case study of the model output in Appendix \ref{sec:diff_backbone} $\sim$ \ref{sec:case_study}, respectively.

\begin{table}[t]
\centering
\small
\setlength\tabcolsep{3pt}
\scalebox{1.0}{
\begin{tabular}{l|c|c|c}
\toprule
\multirow{1}{*}{\textbf{Method}} & \multicolumn{1}{c}{MAVEN}                    & \multicolumn{1}{|c}{ACE05}     & \multicolumn{1}{|c}{Average}                   \\
\midrule

\esco & \textbf{57.35} & \textbf{55.20}  & \textbf{56.28} \\
\emph{w.o.} margin & 55.92  & 54.13  & 55.03 \\
\emph{w.o.} calibration & 55.76  & 53.85  &  54.81 \\
\emph{w.o.} FKT & 56.58  & 54.38  & 55.48  \\
\bottomrule
\end{tabular}
}
\caption{F1 score ($\%$) of different ablations after learning all tasks: (i) without the margin-based loss, (ii) without the memory calibration mechanism, and (iii) without forward knowledge transfer. All components improve the performance of our method.}
\label{table:ablation-brief}
\end{table}

\begin{table}[t]
\centering
\small
\setlength\tabcolsep{3pt}
\scalebox{1.0}
{
\begin{tabular}{l | cc | cc}
    \toprule
    \multirow{2}{*}{\textbf{Dataset}} & \multicolumn{2}{c|} {MAVEN} & \multicolumn{2}{c}{ACE05} \\
    & BWT & FWT & BWT & FWT \\
    \midrule
    EMP & -10.4 & -2.9 & -16.8 & 0.5  \\ 
    \midrule
    \esco & \textbf{-6.3} & \textbf{-1.2} & \textbf{-7.5} & \textbf{4.1} \\
    \bottomrule
    \end{tabular}
} 
\caption{Backward transfer (BWT) and forward transfer (FWT) of EMP and \esco\ after learning all tasks on MAVEN and ACE05.}
\label{tab:knowledge_transfer}
\end{table}
\section{Related Work}

\textbf{Lifelong event detection} (\led) aims to continually learn from a sequence of event detection tasks with different sets of event types. \citet{cao-etal-2020-incremental} propose KCN which addresses the semantic ambiguity and data imbalance problems in \led\ by prototype enhanced retrospection and hierarchical distillation. KT \citep{yu-etal-2021-lifelong} encourages bi-directional knowledge transfer between old and new event types. \citet{liu-etal-2022-incremental} introduce EMP to retain previously learned task-specific event knowledge through soft prompts. In contrast to previous works, we innovate on the methodology by imposing further constraints in the embedding space to mitigate forgetting and overfitting. 
\section{Conclusion}

In this work, we have introduced embedding space separation and compaction (\esco) for lifelong event detection (\led). \esco\ imposes novel feature constraints in the embedding space to alleviate forgetting and overfitting problems. It initializes the learnable parameters for the new task by inheriting those from the previously learned task to facilitate forward knowledge transfer. With extensive experiments and analysis, we have demonstrated that \esco\ significantly outperforms previous methods. For future work, we are interested in exploring \esco\ in a meta-learning paradigm for \led.

\section*{Limitations}
Although effective, \esco\ also has some limitations. For example, \esco\ mainly focuses on the setting where each task has enough training data. Few-shot learning and in-context learning have been receiving increasing interest within the community \citep{ding2024data}. We leave how to explore lifelong event detection in few-shot settings as future work. Besides, as large language models (LLMs) have shown impressive performance on a variety of tasks \citep{brown2020language,ouyang2022training,touvron2023llama,qin2023improving,qin2023context,qin2023chatgpt,achiam2023gpt,cola,ding-etal-2023-gpt}, a further improvement could be to explore lifelong event detection with LLMs.

\bibliography{anthology,custom}

\begin{thebibliography}{35}
\expandafter\ifx\csname natexlab\endcsname\relax\def\natexlab#1{#1}\fi

\bibitem[{Achiam et~al.(2023)Achiam, Adler, Agarwal, Ahmad, Akkaya, Aleman, Almeida, Altenschmidt, Altman, Anadkat et~al.}]{achiam2023gpt}
Josh Achiam, Steven Adler, Sandhini Agarwal, Lama Ahmad, Ilge Akkaya, Florencia~Leoni Aleman, Diogo Almeida, Janko Altenschmidt, Sam Altman, Shyamal Anadkat, et~al. 2023.
\newblock \href {https://arxiv.org/abs/2303.08774} {Gpt-4 technical report}.
\newblock \emph{arXiv preprint arXiv:2303.08774}.

\bibitem[{Brown et~al.(2020)Brown, Mann, Ryder, Subbiah, Kaplan, Dhariwal, Neelakantan, Shyam, Sastry, Askell et~al.}]{brown2020language}
Tom~B Brown, Benjamin Mann, Nick Ryder, Melanie Subbiah, Jared Kaplan, Prafulla Dhariwal, Arvind Neelakantan, Pranav Shyam, Girish Sastry, Amanda Askell, et~al. 2020.
\newblock \href {https://arxiv.org/abs/2005.14165} {Language models are few-shot learners}.
\newblock \emph{arXiv preprint arXiv:2005.14165}.

\bibitem[{Cao et~al.(2020)Cao, Chen, Zhao, and Wang}]{cao-etal-2020-incremental}
Pengfei Cao, Yubo Chen, Jun Zhao, and Taifeng Wang. 2020.
\newblock \href {https://doi.org/10.18653/v1/2020.emnlp-main.52} {Incremental event detection via knowledge consolidation networks}.
\newblock In \emph{Proceedings of the 2020 Conference on Empirical Methods in Natural Language Processing (EMNLP)}, pages 707--717, Online. Association for Computational Linguistics.

\bibitem[{Chen et~al.(2024)Chen, Qin, Jiang, and Choi}]{chen2024large}
Ruirui Chen, Chengwei Qin, Weifeng Jiang, and Dongkyu Choi. 2024.
\newblock \href {https://ojs.aaai.org/index.php/AAAI/article/download/29730/31254} {Is a large language model a good annotator for event extraction?}
\newblock In \emph{Proceedings of the AAAI Conference on Artificial Intelligence}, 16, pages 17772--17780.

\bibitem[{Chen et~al.(2015)Chen, Xu, Liu, Zeng, and Zhao}]{chen-etal-2015-event}
Yubo Chen, Liheng Xu, Kang Liu, Daojian Zeng, and Jun Zhao. 2015.
\newblock \href {https://doi.org/10.3115/v1/P15-1017} {Event extraction via dynamic multi-pooling convolutional neural networks}.
\newblock In \emph{Proceedings of the 53rd Annual Meeting of the Association for Computational Linguistics and the 7th International Joint Conference on Natural Language Processing (Volume 1: Long Papers)}, pages 167--176, Beijing, China. Association for Computational Linguistics.

\bibitem[{Devlin et~al.(2019)Devlin, Chang, Lee, and Toutanova}]{devlin-etal-2019-bert}
Jacob Devlin, Ming-Wei Chang, Kenton Lee, and Kristina Toutanova. 2019.
\newblock \href {https://doi.org/10.18653/v1/N19-1423} {{BERT}: Pre-training of deep bidirectional transformers for language understanding}.
\newblock In \emph{Proceedings of the 2019 Conference of the North {A}merican Chapter of the Association for Computational Linguistics: Human Language Technologies, Volume 1 (Long and Short Papers)}, pages 4171--4186, Minneapolis, Minnesota. Association for Computational Linguistics.

\bibitem[{Ding et~al.(2023)Ding, Qin, Liu, Chia, Li, Joty, and Bing}]{ding-etal-2023-gpt}
Bosheng Ding, Chengwei Qin, Linlin Liu, Yew~Ken Chia, Boyang Li, Shafiq Joty, and Lidong Bing. 2023.
\newblock \href {https://doi.org/10.18653/v1/2023.acl-long.626} {Is {GPT}-3 a good data annotator?}
\newblock In \emph{Proceedings of the 61st Annual Meeting of the Association for Computational Linguistics (Volume 1: Long Papers)}, pages 11173--11195, Toronto, Canada. Association for Computational Linguistics.

\bibitem[{Ding et~al.(2024)Ding, Qin, Zhao, Luo, Li, Chen, Xia, Hu, Luu, and Joty}]{ding2024data}
Bosheng Ding, Chengwei Qin, Ruochen Zhao, Tianze Luo, Xinze Li, Guizhen Chen, Wenhan Xia, Junjie Hu, Anh~Tuan Luu, and Shafiq Joty. 2024.
\newblock \href {https://arxiv.org/abs/2403.02990} {Data augmentation using llms: Data perspectives, learning paradigms and challenges}.
\newblock \emph{arXiv preprint arXiv:2403.02990}.

\bibitem[{Doddington et~al.(2004)Doddington, Mitchell, Przybocki, Ramshaw, Strassel, and Weischedel}]{doddington-etal-2004-automatic}
George Doddington, Alexis Mitchell, Mark Przybocki, Lance Ramshaw, Stephanie Strassel, and Ralph Weischedel. 2004.
\newblock \href {http://www.lrec-conf.org/proceedings/lrec2004/pdf/5.pdf} {The automatic content extraction ({ACE}) program {--} tasks, data, and evaluation}.
\newblock In \emph{Proceedings of the Fourth International Conference on Language Resources and Evaluation ({LREC}{'}04)}, Lisbon, Portugal. European Language Resources Association (ELRA).

\bibitem[{Han et~al.(2020)Han, Dai, Gao, Lin, Liu, Li, Sun, and Zhou}]{han-etal-2020-continual}
Xu~Han, Yi~Dai, Tianyu Gao, Yankai Lin, Zhiyuan Liu, Peng Li, Maosong Sun, and Jie Zhou. 2020.
\newblock \href {https://doi.org/10.18653/v1/2020.acl-main.573} {Continual relation learning via episodic memory activation and reconsolidation}.
\newblock In \emph{Proceedings of the 58th Annual Meeting of the Association for Computational Linguistics}, pages 6429--6440, Online. Association for Computational Linguistics.

\bibitem[{Huang and Ji(2020)}]{huang-ji-2020-semi}
Lifu Huang and Heng Ji. 2020.
\newblock \href {https://doi.org/10.18653/v1/2020.emnlp-main.53} {Semi-supervised new event type induction and event detection}.
\newblock In \emph{Proceedings of the 2020 Conference on Empirical Methods in Natural Language Processing (EMNLP)}, pages 718--724, Online. Association for Computational Linguistics.

\bibitem[{Ke et~al.(2020)Ke, Liu, and Huang}]{ke2020continual}
Zixuan Ke, Bing Liu, and Xingchang Huang. 2020.
\newblock \href {https://proceedings.neurips.cc/paper/2020/file/d7488039246a405baf6a7cbc3613a56f-Paper.pdf} {Continual learning of a mixed sequence of similar and dissimilar tasks}.
\newblock In \emph{Advances in Neural Information Processing Systems}, volume~33, pages 18493--18504. Curran Associates, Inc.

\bibitem[{Liu et~al.(2022)Liu, Chang, and Huang}]{liu-etal-2022-incremental}
Minqian Liu, Shiyu Chang, and Lifu Huang. 2022.
\newblock \href {https://aclanthology.org/2022.coling-1.189} {Incremental prompting: Episodic memory prompt for lifelong event detection}.
\newblock In \emph{Proceedings of the 29th International Conference on Computational Linguistics}, pages 2157--2165, Gyeongju, Republic of Korea. International Committee on Computational Linguistics.

\bibitem[{Liu et~al.(2019)Liu, Ott, Goyal, Du, Joshi, Chen, Levy, Lewis, Zettlemoyer, and Stoyanov}]{liu2019roberta}
Yinhan Liu, Myle Ott, Naman Goyal, Jingfei Du, Mandar Joshi, Danqi Chen, Omer Levy, Mike Lewis, Luke Zettlemoyer, and Veselin Stoyanov. 2019.
\newblock \href {https://arxiv.org/abs/1907.11692} {Roberta: A robustly optimized bert pretraining approach}.
\newblock \emph{arXiv preprint arXiv:1907.11692}.

\bibitem[{Lopez-Paz and Ranzato(2017)}]{lopez2017gradient}
David Lopez-Paz and Marc'Aurelio Ranzato. 2017.
\newblock \href {https://arxiv.org/abs/1706.08840} {Gradient episodic memory for continual learning}.
\newblock \emph{Advances in neural information processing systems}, 30.

\bibitem[{Lu et~al.(2022)Lu, Xia, Arora, and Hazan}]{samuel}
Zhou Lu, Wenhan Xia, Sanjeev Arora, and Elad Hazan. 2022.
\newblock \href {https://arxiv.org/abs/2203.01400} {Adaptive gradient methods with local guarantees}.
\newblock \emph{arXiv preprint arXiv:2203.01400}.

\bibitem[{McCloskey and Cohen(1989)}]{mccloskey1989catastrophic}
Michael McCloskey and Neal~J Cohen. 1989.
\newblock \href {https://doi.org/https://doi.org/10.1016/S0079-7421(08)60536-8} {Catastrophic interference in connectionist networks: The sequential learning problem}.
\newblock In \emph{Psychology of learning and motivation}, volume~24, pages 109--165. Elsevier.

\bibitem[{Nguyen et~al.(2016)Nguyen, Cho, and Grishman}]{nguyen-etal-2016-joint-event}
Thien~Huu Nguyen, Kyunghyun Cho, and Ralph Grishman. 2016.
\newblock \href {https://doi.org/10.18653/v1/N16-1034} {Joint event extraction via recurrent neural networks}.
\newblock In \emph{Proceedings of the 2016 Conference of the North {A}merican Chapter of the Association for Computational Linguistics: Human Language Technologies}, pages 300--309, San Diego, California. Association for Computational Linguistics.

\bibitem[{Ouyang et~al.(2022)Ouyang, Wu, Jiang, Almeida, Wainwright, Mishkin, Zhang, Agarwal, Slama, Ray et~al.}]{ouyang2022training}
Long Ouyang, Jeffrey Wu, Xu~Jiang, Diogo Almeida, Carroll Wainwright, Pamela Mishkin, Chong Zhang, Sandhini Agarwal, Katarina Slama, Alex Ray, et~al. 2022.
\newblock \href {https://arxiv.org/abs/2203.02155} {Training language models to follow instructions with human feedback}.
\newblock \emph{Advances in Neural Information Processing Systems}, 35:27730--27744.

\bibitem[{Qin and Joty(2022{\natexlab{a}})}]{qin-joty-2022-continual}
Chengwei Qin and Shafiq Joty. 2022{\natexlab{a}}.
\newblock \href {https://doi.org/10.18653/v1/2022.acl-long.198} {Continual few-shot relation learning via embedding space regularization and data augmentation}.
\newblock In \emph{Proceedings of the 60th Annual Meeting of the Association for Computational Linguistics (Volume 1: Long Papers)}, pages 2776--2789, Dublin, Ireland. Association for Computational Linguistics.

\bibitem[{Qin and Joty(2022{\natexlab{b}})}]{qin2022lfpt}
Chengwei Qin and Shafiq Joty. 2022{\natexlab{b}}.
\newblock \href {https://openreview.net/forum?id=HCRVf71PMF} {{LFPT}5: A unified framework for lifelong few-shot language learning based on prompt tuning of t5}.
\newblock In \emph{International Conference on Learning Representations}.

\bibitem[{Qin et~al.(2023{\natexlab{a}})Qin, Joty, and Chen}]{qin2023lifelong}
Chengwei Qin, Shafiq Joty, and Chen Chen. 2023{\natexlab{a}}.
\newblock \href {https://arxiv.org/abs/2310.09886} {Lifelong sequence generation with dynamic module expansion and adaptation}.
\newblock \emph{arXiv preprint arXiv:2310.09886}.

\bibitem[{Qin et~al.(2023{\natexlab{b}})Qin, Joty, Li, and Zhao}]{qin-etal-2023-learning}
Chengwei Qin, Shafiq Joty, Qian Li, and Ruochen Zhao. 2023{\natexlab{b}}.
\newblock \href {https://doi.org/10.18653/v1/2023.acl-long.659} {Learning to initialize: Can meta learning improve cross-task generalization in prompt tuning?}
\newblock In \emph{Proceedings of the 61st Annual Meeting of the Association for Computational Linguistics (Volume 1: Long Papers)}, pages 11802--11832, Toronto, Canada. Association for Computational Linguistics.

\bibitem[{Qin et~al.(2023{\natexlab{c}})Qin, Xia, Jiao, and Joty}]{qin2023improving}
Chengwei Qin, Wenhan Xia, Fangkai Jiao, and Shafiq Joty. 2023{\natexlab{c}}.
\newblock \href {https://arxiv.org/abs/2312.17055} {Improving in-context learning via bidirectional alignment}.
\newblock \emph{arXiv preprint arXiv:2312.17055}.

\bibitem[{Qin et~al.(2023{\natexlab{d}})Qin, Zhang, Dagar, and Ye}]{qin2023context}
Chengwei Qin, Aston Zhang, Anirudh Dagar, and Wenming Ye. 2023{\natexlab{d}}.
\newblock \href {https://arxiv.org/abs/2310.09881} {In-context learning with iterative demonstration selection}.
\newblock \emph{arXiv preprint arXiv:2310.09881}.

\bibitem[{Qin et~al.(2023{\natexlab{e}})Qin, Zhang, Zhang, Chen, Yasunaga, and Yang}]{qin2023chatgpt}
Chengwei Qin, Aston Zhang, Zhuosheng Zhang, Jiaao Chen, Michihiro Yasunaga, and Diyi Yang. 2023{\natexlab{e}}.
\newblock \href {https://arxiv.org/abs/2302.06476} {Is chatgpt a general-purpose natural language processing task solver?}
\newblock \emph{arXiv preprint arXiv:2302.06476}.

\bibitem[{Sun et~al.(2022)Sun, Lyu, Shang, Feng, and Wan}]{sun2022exploring}
Qing Sun, Fan Lyu, Fanhua Shang, Wei Feng, and Liang Wan. 2022.
\newblock \href {https://openreview.net/forum?id=u4dXcUEsN7B} {Exploring example influence in continual learning}.
\newblock In \emph{Advances in Neural Information Processing Systems}.

\bibitem[{Touvron et~al.(2023)Touvron, Martin, Stone, Albert, Almahairi, Babaei, Bashlykov, Batra, Bhargava, Bhosale et~al.}]{touvron2023llama}
Hugo Touvron, Louis Martin, Kevin Stone, Peter Albert, Amjad Almahairi, Yasmine Babaei, Nikolay Bashlykov, Soumya Batra, Prajjwal Bhargava, Shruti Bhosale, et~al. 2023.
\newblock \href {https://arxiv.org/abs/2307.09288} {Llama 2: Open foundation and fine-tuned chat models}.
\newblock \emph{arXiv preprint arXiv:2307.09288}.

\bibitem[{Van~der Maaten and Hinton(2008)}]{van2008visualizing}
Laurens Van~der Maaten and Geoffrey Hinton. 2008.
\newblock \href {https://www.jmlr.org/papers/v9/vandermaaten08a.html} {Visualizing data using t-sne.}
\newblock \emph{Journal of machine learning research}, 9(11).

\bibitem[{Wang et~al.(2020)Wang, Wang, Han, Jiang, Han, Liu, Li, Li, Lin, and Zhou}]{wang-etal-2020-maven}
Xiaozhi Wang, Ziqi Wang, Xu~Han, Wangyi Jiang, Rong Han, Zhiyuan Liu, Juanzi Li, Peng Li, Yankai Lin, and Jie Zhou. 2020.
\newblock \href {https://doi.org/10.18653/v1/2020.emnlp-main.129} {{MAVEN}: {A} {M}assive {G}eneral {D}omain {E}vent {D}etection {D}ataset}.
\newblock In \emph{Proceedings of the 2020 Conference on Empirical Methods in Natural Language Processing (EMNLP)}, pages 1652--1671, Online. Association for Computational Linguistics.

\bibitem[{Welling(2009)}]{welling2009herding}
Max Welling. 2009.
\newblock \href {http://machinelearning.org/archive/icml2009/papers/447.pdf} {Herding dynamical weights to learn}.
\newblock In \emph{Proceedings of the 26th Annual International Conference on Machine Learning}, pages 1121--1128.

\bibitem[{Wolf et~al.(2020)Wolf, Debut, Sanh, Chaumond, Delangue, Moi, Cistac, Rault, Louf, Funtowicz, Davison, Shleifer, von Platen, Ma, Jernite, Plu, Xu, Le~Scao, Gugger, Drame, Lhoest, and Rush}]{wolf-etal-2020-transformers}
Thomas Wolf, Lysandre Debut, Victor Sanh, Julien Chaumond, Clement Delangue, Anthony Moi, Pierric Cistac, Tim Rault, Remi Louf, Morgan Funtowicz, Joe Davison, Sam Shleifer, Patrick von Platen, Clara Ma, Yacine Jernite, Julien Plu, Canwen Xu, Teven Le~Scao, Sylvain Gugger, Mariama Drame, Quentin Lhoest, and Alexander Rush. 2020.
\newblock \href {https://doi.org/10.18653/v1/2020.emnlp-demos.6} {Transformers: State-of-the-art natural language processing}.
\newblock In \emph{Proceedings of the 2020 Conference on Empirical Methods in Natural Language Processing: System Demonstrations}, pages 38--45, Online. Association for Computational Linguistics.

\bibitem[{Wu et~al.(2019)Wu, Chen, Wang, Ye, Liu, Guo, and Fu}]{wu2019large}
Yue Wu, Yinpeng Chen, Lijuan Wang, Yuancheng Ye, Zicheng Liu, Yandong Guo, and Yun Fu. 2019.
\newblock \href {https://arxiv.org/abs/1905.13260} {Large scale incremental learning}.
\newblock In \emph{Proceedings of the IEEE/CVF Conference on Computer Vision and Pattern Recognition}, pages 374--382.

\bibitem[{Xia et~al.(2024)Xia, Qin, and Hazan}]{cola}
Wenhan Xia, Chengwei Qin, and Elad Hazan. 2024.
\newblock Chain of lora: Efficient fine-tuning of language models via residual learning.
\newblock \emph{arXiv preprint arXiv:2401.04151}.

\bibitem[{Yu et~al.(2021)Yu, Ji, and Natarajan}]{yu-etal-2021-lifelong}
Pengfei Yu, Heng Ji, and Prem Natarajan. 2021.
\newblock \href {https://doi.org/10.18653/v1/2021.emnlp-main.428} {Lifelong event detection with knowledge transfer}.
\newblock In \emph{Proceedings of the 2021 Conference on Empirical Methods in Natural Language Processing}, pages 5278--5290, Online and Punta Cana, Dominican Republic. Association for Computational Linguistics.

\end{thebibliography}

\appendix

\section{Appendix}
\label{sec:appendix}

\begin{table}[t]
\centering
\small
\setlength\tabcolsep{3pt}
\scalebox{1.0}{
\begin{tabular}{l | ccccc}
    \toprule
    & \multicolumn{5}{c}{MAVEN} \\
    \midrule
    Task index &1 &2 & 3 & 4 & 5 \\ 
    \midrule
    EMP & \textbf{67.1} & 58.3 & 55.7 & 53.2 & 52.9  \\ 
    \esco & \textbf{67.1} & \textbf{60.8} & \textbf{59.0} & \textbf{55.3} & \textbf{55.1}  \\ 
    \bottomrule
    \end{tabular}
    } 
    \caption{Performance comparison between EMP and \esco\ on MAVEN using RoBERTa as the backbone.}
    \label{tab:different_backbone_models}
\end{table}

\begin{table*}[!t]
	\centering
	\resizebox{0.9\textwidth}{!}
	{
	\begin{tabular}{l | ccccc | ccccc}
    \toprule
    & \multicolumn{5}{c}{MAVEN}  & \multicolumn{5}{c}{ACE05} \\
    \midrule
    Task index &1 &2 & 3 & 4 & 5 & 1 &2 & 3 & 4 & 5  \\ 
    \midrule
    EMP & $\mathbf{67.50}_{\pm 3.54}$ & $59.67_{\pm 2.74}$ & $58.03_{\pm 1.44}$ & $54.80_{\pm 0.95}$ & $54.39_{\pm 0.82}$ & $\mathbf{58.35}_{\pm 6.92}$ & $50.03_{\pm 18.18}$ & $54.91_{\pm 9.19}$ & $47.78_{\pm 2.57}$ & $47.19_{\pm 8.53}$ \\ 
    \esco & $\mathbf{67.50}_{\pm 3.54}$ & $\mathbf{61.37}_{\pm 2.92}$ & $\mathbf{60.65}_{\pm 1.85}$ & $\mathbf{57.43}_{\pm 0.81}$ & $\mathbf{57.35}_{\pm 0.66}$ & $\mathbf{58.35}_{\pm 6.92}$ & $\mathbf{57.42}_{\pm 13.56}$ & $\mathbf{57.63}_{\pm 6.26}$ & $\mathbf{53.64}_{\pm 4.41}$ & $\mathbf{55.20}_{\pm 4.16}$ \\ 
    \bottomrule
    \end{tabular}
    } 
    \caption{F1 score ($\%$) and variance of EMP and \esco\ at every time step on two datasets.}
    \label{tab:result_with_variance}
\end{table*}

\begin{table*}[!t]
	\centering
	\resizebox{0.9\textwidth}{!}
	{
	\begin{tabular}{ll | ccccc | ccccc}
    \toprule
    \multicolumn{2}{c}{\multirow{2}*{\textbf{Task Order}}} & \multicolumn{5}{c}{MAVEN}  & \multicolumn{5}{c}{ACE05} \\
    \cmidrule{3-12}
    \multicolumn{2}{c}{} &1 &2 & 3 & 4 & 5 & 1 &2 & 3 & 4 & 5  \\ 
    \midrule
    \multicolumn{2}{c}{\multirow{2}*{{1}}} & \textbf{70.80} & 62.19 & 60.00 & 56.11 & 54.95 & \textbf{62.58} & 65.60 & \textbf{67.20} & 45.06 & 39.07 \\ 
    \multicolumn{2}{c}{} & \textbf{70.80} & \textbf{64.73} & \textbf{62.22} & \textbf{57.26} & \textbf{57.67} & \textbf{62.58} & \textbf{68.35} & 66.02 & \textbf{54.13} & \textbf{49.33} \\
    \midrule
    \multicolumn{2}{c}{\multirow{2}*{{2}}} & \textbf{66.06} & 56.01 & 56.16	& 54.07 & 54.91 & \textbf{49.17} &	50.99 &	57.78	& 51.76	& 42.49 \\ 
    \multicolumn{2}{c}{} & \textbf{66.06}&	\textbf{58.36}	&\textbf{58.16}	&\textbf{57.71}	&\textbf{57.40}&\textbf{49.17}	&\textbf{52.13}	&\textbf{60.73}	&\textbf{57.85}	&\textbf{54.76} \\ 
    \midrule
    \multicolumn{2}{c}{\multirow{2}*{{3}}} & \textbf{70.80}	&59.91	&57.43	&55.53	&55.02 & \textbf{62.58}	&59.43	&50.08	&46.27	&55.12\\ 
    \multicolumn{2}{c}{} & \textbf{70.80}	&\textbf{62.70}	&\textbf{60.03}	&\textbf{58.27}	&\textbf{57.57} & \textbf{62.58}	&\textbf{61.03}	&\textbf{51.96}	&\textbf{47.95}	&\textbf{59.97}
    \\ 
    \midrule
    \multicolumn{2}{c}{\multirow{2}*{{4}}} & \textbf{67.42}	&62.35	&58.74	&54.10	&53.19 & \textbf{64.68}	&55.15	&56.87	&47.30	&41.65\\ 
    \multicolumn{2}{c}{} & \textbf{67.42}	&\textbf{62.81}	&\textbf{62.72}	&\textbf{56.12}	&\textbf{56.20} & \textbf{64.68}	& \textbf{69.12}	& \textbf{58.44}	& \textbf{57.78}	& \textbf{53.67} \\ 
    \midrule
    \multicolumn{2}{c}{\multirow{2}*{{5}}} & \textbf{62.39}	&57.90	&57.81	&54.20	&53.88 &\textbf{52.74}	&18.99	&42.62	&48.53	&57.64 \\ 
    \multicolumn{2}{c}{} & \textbf{62.39}	&\textbf{58.22}	&\textbf{60.09}	&\textbf{57.78}	&\textbf{57.88} & \textbf{52.74}	&\textbf{36.49}	&\textbf{51.01}	&\textbf{50.46}	&\textbf{58.29} \\ 
    \bottomrule
    \end{tabular}
    } 
    \caption{F1 score ($\%$) of 5 runs with different task orders  on two datasets. For every order, the upper row shows the performance of EMP and the lower row is the result of \esco.}
    \label{tab:detailed_results_all_orders}
\end{table*}

\begin{table}[t]
\centering
\small
\setlength\tabcolsep{3pt}
\scalebox{1.0}{
\begin{tabular}{l|c|c}
\toprule
\multirow{1}{*}{} & \multicolumn{1}{c}{EMP}                    & \multicolumn{1}{|c}{\esco}                     \\
\midrule
Herding algorithm &  54.4 &  \textbf{57.1} \\
Example influence &  53.5 &  \textbf{56.4} \\
\bottomrule
\end{tabular}
}
\caption{F1 score ($\%$) of EMP and \esco\ with different memory sample selection approaches.}
\label{table:different_sample_selection}
\end{table}

\begin{table}[t]
\centering
\small
\setlength\tabcolsep{3pt}
\scalebox{1.0}{
\begin{tabular}{l|c|c}
\toprule
\multirow{1}{*}{} & \multicolumn{1}{c}{\esco}                    & \multicolumn{1}{|c}{$\esco_{\text{con}}$}                     \\
\midrule
F1 score ($\%$) &  \textbf{57.1}  & 56.6 \\
\bottomrule
\end{tabular}
}
\caption{Performance comparison between \esco\ and $\esco_{\text{con}}$.}
\label{table:comparision_with_contrastive}
\end{table}

\begin{table*}[t]
\centering
\begin{tabular}{ll}
\toprule
\multicolumn{2}{l}{Fighting continued until 9 May, when the Red Army \blue{entered} the nearly liberated city.}                                        \\ \midrule
\multicolumn{1}{l}{\textbf{Label}}     & \textit{Arriving}\\ \hline
\multicolumn{1}{l}{\textbf{EMP}}       & \textit{\red{Becoming\_a\_member}}             \\ \hline
\multicolumn{1}{l}{\textbf{\esco}}     & \textit{\textcolor{applegreen}{Arriving}}                                                                                   \\ \toprule
\multicolumn{2}{l}{In Japan, hundreds of people \blue{evacuated} from mudslide-prone areas.}                                                              \\ \midrule
\multicolumn{1}{l}{\textbf{Label}}     & \textit{Escaping}                       \\ \hline
\multicolumn{1}{l}{\textbf{EMP}}    & \textit{\red{Removing}}                                                                                                                     \\ \hline
\multicolumn{1}{l}{\textbf{\esco}}          & \textit{\textcolor{applegreen}{Escaping}}                                                                                                \\ \bottomrule
\end{tabular}
\caption{Output examples of EMP and \esco. We color target spans in \blue{blue}, correct outputs in \textcolor{applegreen}{green}, and wrong outputs in \red{red}.}
\label{table:case-study-example}
\end{table*}

\subsection{Overlap of Feature Distributions} \label{sec:overlap_distribution}

\begin{figure}[t]
    \centering
    \includegraphics[width=0.48\textwidth]{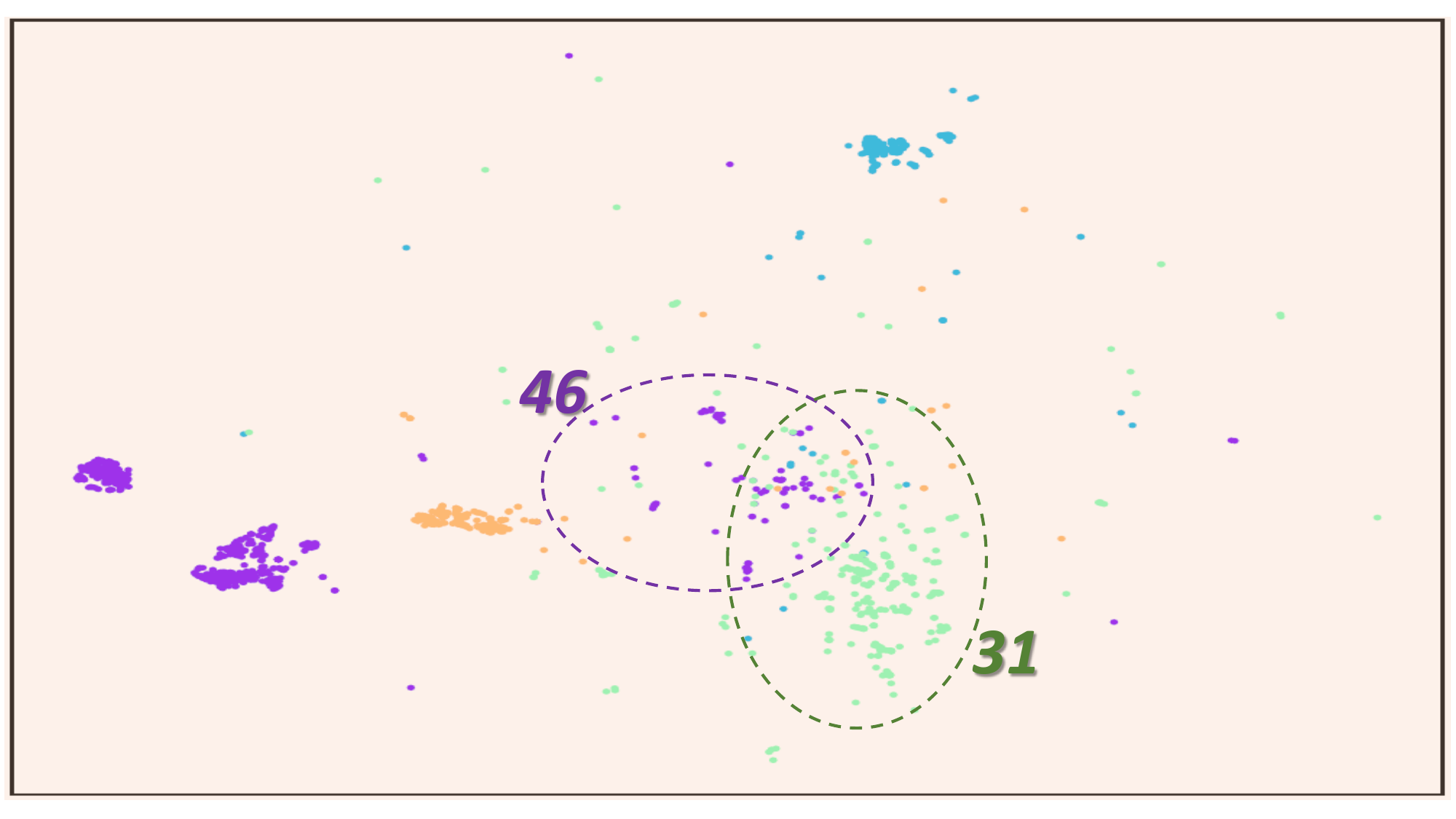}
    \caption{Overlap between feature distributions of event types at different learning stages, \eg\ circled regions. Colors represent different event types with numbers being the event indexes.}
    \label{fig:distribution_overlap}
\end{figure}

\Cref{fig:distribution_overlap} shows an example of the overlap between feature distributions of event types at different learning stages.

\subsection{Implementation Details} \label{sec:exp_details}

All methods are implemented with PyTorch/Transformers library \citep{wolf-etal-2020-transformers}. For hyperparameters, we mainly follow the settings in \citet{liu-etal-2022-incremental} to have a fair comparison. We adopt $-0.1$ for the margin value $m_1$ so that the similarity score between a new sample and the prototype of a previously learned event type could be optimized to a negative number, \ie\ large inter-class distance. We set the weight $\lambda_{1}$ to $0.1$ so that its corresponding loss $\mathcal{L}_{\text{sim}}$ has roughly the same order of magnitude as other losses. The loss weight $\lambda_{2}$ is set to $\frac{s}{k+s}$, where $k$ is the number of target spans in the current batch and $s$ is equal to $50$ following \citet{liu-etal-2022-incremental}. For each task, we train the model for 20 epochs with early stopping. 

For the state-of-the-art EMP \citep{liu-etal-2022-incremental}, we reproduce the results using its open-source code and the same environment. For our method, we use the same environment and shared hyperparameters as EMP. For other baselines, we reuse the results in \citet{liu-etal-2022-incremental}. There are mainly two reasons: \Na They perform much worse than EMP, \ie\ they are not primary comparison approaches in our work; and \Nb EMP reports different baseline results from \citet{yu-etal-2021-lifelong}, indicating different settings. However, EMP does not provide details on how to obtain baseline results. As we use the same setting as EMP, we decide to reuse its results for other baselines.

\subsection{Results with Variance} \label{sec:results_with_variance}

We show results with variance for EMP and \esco\ in \Cref{tab:result_with_variance}.

\subsection{Detailed Results for Different Task Orders} \label{sec:results_different_oders}

\Cref{tab:detailed_results_all_orders} reports detailed results for different task orders.

\begin{figure}[t]
    \centering
    \includegraphics[width=0.48\textwidth]{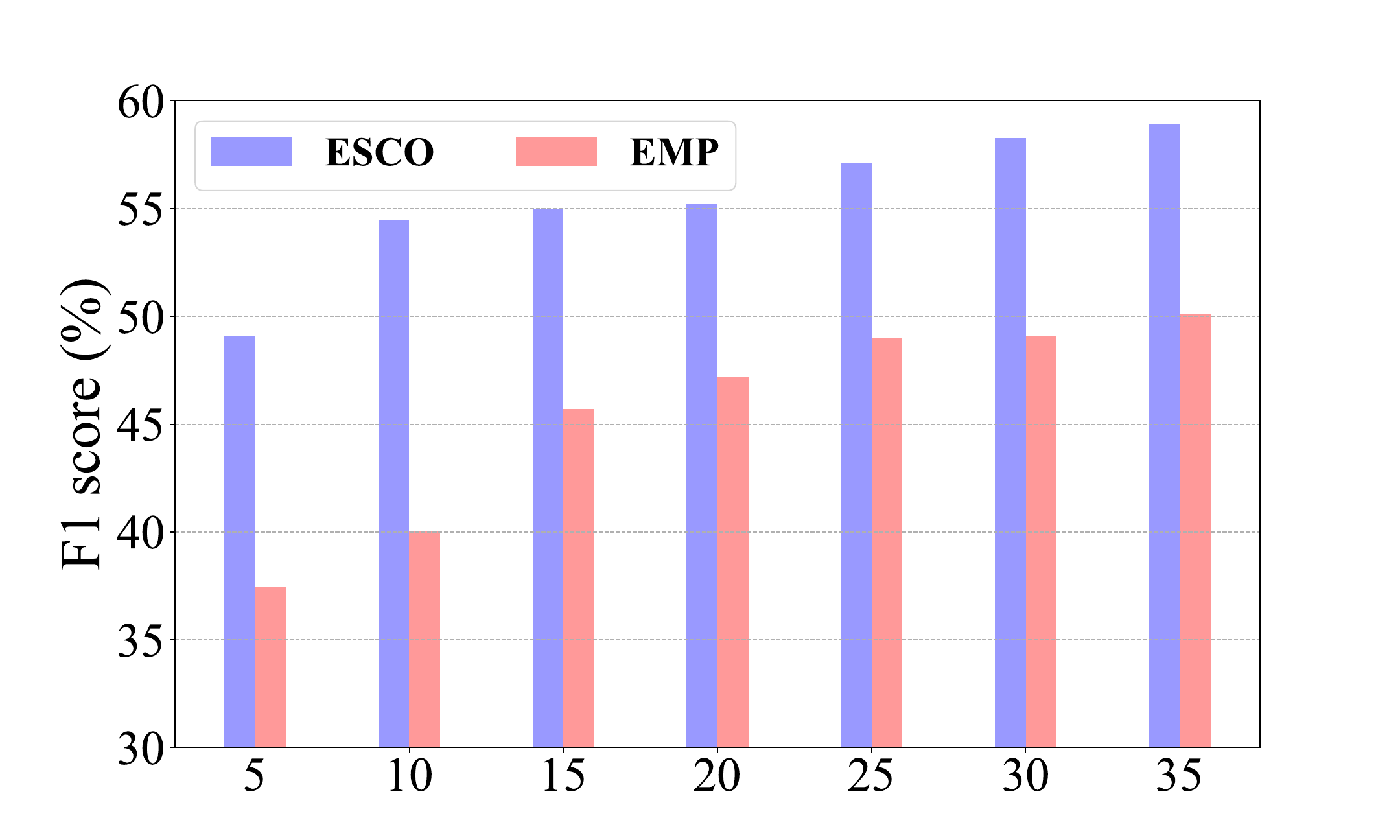}
    \caption{The performance of \esco\ and EMP with different memory sizes.}
    \label{fig:diff_mem}
\end{figure}

\subsection{Different Backbone Model} \label{sec:diff_backbone}
To investigate the generalization ability of \esco, we further conduct experiments on MAVEN using RoBERTa \citep{liu2019roberta} backbone. The F1 scores of EMP and ESCO at each time step are reported in \Cref{tab:different_backbone_models}, which verify that \esco\ can indeed generalize to different models.

\subsection{Effect of Memory Size} \label{sec:memory_size}

Following \citet{liu-etal-2022-incremental}, we select $20$ samples as memory data for each event type. To investigate whether different memory sizes influence the performance gain of \esco, we conduct controlled experiments on ACE05 with memory size $\{5,10,15,25,30,35\}$. The performance comparison between \esco\ and EMP is shown in \Cref{fig:diff_mem}. We can observe that \esco\ consistently outperforms EMP by a large margin with different memory sizes, demonstrating its robustness.

\subsection{Different Memory Sample Selection Approaches} \label{sec:diff_memory_selection}
Following EMP \citep{liu-etal-2022-incremental}, we use the herding algorithm \citep{welling2009herding} to select memory samples. To validate whether different memory sample selection approaches influence the performance gain of \esco, we replace the herding algorithm of EMP and \esco\ with \emph{example influence} \citep{sun2022exploring} for memory selection. We randomly select three sequences for experiments and report the performance comparison between EMP and \esco\ in \Cref{table:different_sample_selection}. We can see that: \Na \esco\ consistently outperforms EMP in different cases, demonstrating its effectiveness; and \Nb herding algorithm performs better than example influence, justifying our choice. 

\subsection{Comparison with the Contrastive Loss} \label{sec:comparison_contrastive}

As mentioned in \Cref{sec:methodology}, our designed margin-based loss $\mathcal{L}_{\text{sim}}$ is different from the contrastive loss as $\mathcal{L}_{\text{sim}}$ only includes negative pairs while ignoring positive ones. To further demonstrate its superiority, we replace it with the contrastive loss in \citep{qin-joty-2022-continual}, namely $\esco_{\text{con}}$. We use the same sequences as \Cref{sec:diff_memory_selection} for experiments and report the results of \esco\ and $\esco_{\text{con}}$ in \Cref{table:comparision_with_contrastive}, which verify the effectiveness of $\mathcal{L}_{\text{sim}}$.

\subsection{Case Study} \label{sec:case_study}

We select MAVEN as a representative task and show several example outputs in \Cref{table:case-study-example}. Compared with EMP, \esco\ is able to retain more precise and fine-grained event knowledge, \eg\ \esco\ can successfully detect the event type \emph{Escaping} from the target span \emph{evacuated} while EMP is confused by another semantically similar type \emph{Removing}.

\end{document}